\documentclass[10pt,twocolumn,letterpaper]{article}

\usepackage[pagenumbers]{cvpr} 

\usepackage{graphicx}
\usepackage{amsmath}
\usepackage{amssymb}
\usepackage{booktabs}
\usepackage{tabulary,multirow,overpic,xcolor}
\usepackage{algorithm}
\usepackage{algpseudocode}
\usepackage{listings}
\usepackage{color}
\usepackage{xcolor,colortbl}
\usepackage{setspace}

\usepackage[pagebackref,breaklinks,colorlinks]{hyperref}

\usepackage[capitalize]{cleveref}
\crefname{section}{Sec.}{Secs.}
\crefname{section}{Section}{Sections}
\crefname{table}{Table}{Tables}
\crefname{table}{Tab.}{Tabs.}

\def\cvprPaperID{10172} 
\def\confName{CVPR}
\def\confYear{2023}

\begin{document}

\title{MAGVLT: Masked Generative Vision-and-Language Transformer}


\author{Sungwoong Kim$^{1,*,\dagger}$,\qquad Daejin Jo$^{2,*}$,\qquad Donghoon Lee$^{2,*}$,\qquad Jongmin Kim$^{2,*}$\\
$^{1}$Department of Artificial Intelligence, Korea University, Seoul, South Korea\\  $^{2}$Kakao Brain, Seongnam, South Korea\\
{\tt\small swkim01@korea.ac.kr, \{daejin.jo, dhlee, jmkim\}@kakaobrain.com}
}
\maketitle
\def\thefootnote{$*$}\footnotetext{Contributed equally.}
\def\thefootnote{$\dagger$}\footnotetext{Work done at Kakao Brain. Corresponding author.}

\begin{abstract}
While generative modeling on multimodal image-text data has been actively developed with large-scale paired datasets, there have been limited attempts to generate both image and text data by a single model rather than a generation of one fixed modality conditioned on the other modality. In this paper, we explore a unified generative vision-and-language (VL) model that can produce both images and text sequences. Especially, we propose a generative VL transformer based on the non-autoregressive mask prediction, named {\bf MAGVLT}, and compare it with an autoregressive generative VL transformer (ARGVLT). In comparison to ARGVLT, the proposed MAGVLT enables bidirectional context encoding, fast decoding by parallel token predictions in an iterative refinement, and extended editing capabilities such as image and text infilling. For rigorous training of our MAGVLT with image-text pairs from scratch, we combine the image-to-text, text-to-image, and joint image-and-text mask prediction tasks. Moreover, we devise two additional tasks based on the step-unrolled mask prediction and the selective prediction on the mixture of two image-text pairs. Experimental results on various downstream generation tasks of VL benchmarks show that our MAGVLT outperforms ARGVLT by a large margin even with significant inference speedup. Particularly, MAGVLT achieves competitive results on both zero-shot image-to-text and text-to-image generation tasks from MS-COCO by one moderate-sized model (fewer than 500M parameters) even without the use of monomodal data and networks.
\end{abstract}
\vspace{-0.5cm}

\section{Introduction}
\label{sec:intro}

Generalizable multimodal modeling has recently made a lot of progress, especially in the field of vision-and-language (VL) modeling \cite{clip, align, dalle, dalle2, imagen, parti, beit3, ding2022cogview2, alayrac2022flamingo, coca, ofa}. In particular, a massive amount of image-text data \cite{wit, cc3m, cc12m, LAION-400M, LAION-5B, coyo-700m} allows robust pretraining of large-scale multimodal VL models that can be easily transferred to various downstream tasks including image captioning \cite{coco, nocaps}, text-guided image generation \cite{nichol2021glide, dalle, dalle2, ding2021cogview, ding2022cogview2, imagen, parti, ERNIE-ViLG2}, visual question answering \cite{vqa2017}, and image-text retrieval \cite{coco, clip, align, flickr30k}. In this respect, many multimodal VL pretraining algorithms have been proposed in the literature, and these can be broadly categorized into either discriminative or generative learning algorithms. Discriminative pretraining such as contrastive learning \cite{clip, align} aims to obtain semantic representations effective for discriminative tasks while generative pretraining learns them to reconstruct an input \cite{wang2021simvlm, vl-beit, beit3, git, anonymous2023connecting, kim2022verse, hao2022language, maskvlm, vl-t5}. The recent growth of model capacity and data size has led to more interest in generative pretraining since it can provide more diverse and improved generalization ability both for VL understanding and VL generation tasks.

While generative VL pretraining has been widely exploited, most existing algorithms focus on representation learning for VL understanding tasks \cite{vl-beit, beit3, maskvlm, vilt, vlmo, uniter, vl-t5} or conditional generation tasks where a generation is performed on one fixed modality conditioned on the other modality \cite{nichol2021glide, dalle, dalle2, ding2021cogview, ding2022cogview2, imagen, parti, ERNIE-ViLG2, wang2021simvlm, git, hao2022language, vl-t5, dai2022enabling, jin2021good, alayrac2022flamingo, coca}. A few algorithms have tried to produce data in both modalities from a single VL model \cite{anonymous2023connecting, kim2022verse}. If one universal model can generate both modalities, it would be beneficial in concentrating training efforts on a single model as well as resource-saving under a resource-constrained deployment. Moreover, we can expect task extension as well as synergetic performance improvement between the modalities from this multimodal generation. Therefore, in this work, we develop a unified generative VL model.

There are two prevalent paradigms of the generative modeling for image and text processing: autoregressive (AR) generative modeling \cite{dalle, ding2021cogview, kim2022verse, parti, git} and non-AR generative modeling \cite{dalle2, imagen, d3pm, anonymous2023connecting}. Many previous algorithms adopt AR modeling due to its excellent generation results and high training scalability, especially with transformer networks. However, AR modeling has restrictions in unidirectional conditioning in that an image needs to be flattened into a 1D sequence by an unnatural ordering. In addition, AR sampling is performed by one-by-one predictions of elements, which incurs very slow generation for a long sequence. Recently, in order to overcome these limitations of AR modeling, non-AR generative modeling based on the mask prediction has been proposed for language \cite{ghazvininejad-etal-2019-mask}, image \cite{chang2022maskgit, Draft-and-Revise}, and video processing \cite{maskvit, phenaki}. Masked modeling is usually employed for representation learning to solve understanding tasks in language, vision, and VL domains. However, with an iterative refinement-based generation and a variable mask ratio during training, it has been shown to be used as a promising generative modeling. In this regard, for our generative VL modeling, we propose {\bf Ma}sked {\bf G}enerative {\bf VL T}ransformer (MAGVLT). In contrast to AR-based generative VL transformer (ARGVLT), the proposed MAGVLT is able to exploit bidirectional conditioning and fast generation through a small number of refinement steps and parallel token predictions.

In specific, MAGVLT can generate any or both of an image and a text sequence conditioned also on any or both of them. Namely, it can perform any kind of task in a form of image-and-text-to-image-and-text (IT2IT), including image-to-text (I2T) and text-to-image (T2I) tasks. Following the previous masked generative modeling \cite{ghazvininejad-etal-2019-mask, chang2022maskgit}, we conduct sampling by iterative denoising based on the masked token prediction and train MAGVLT by the masked token prediction objective with a randomly sampled mask ratio to take into account various denoising steps. Here, to perform robust training of MAGVLT especially with only image-text pairs from scratch, MAGVLT is learned basically by the composition of image-to-text, text-to-image, and joint image-and-text mask prediction objectives. We observe that our cross-modal masking (joint image-and-text mask prediction) during training helps in improving both performances of I2T and T2I tasks over single-modal masking (image-to-text + text-to-image mask predictions). Note that only masked generative modeling used in MAGVLT enables this cross-modal mask prediction during training.

In addition, we propose to use two additional tasks based on the step-unrolled mask prediction and the selective prediction on the mixture of two image-text pairs. The former one is motivated by SUNDAE \cite{savinov2022stepunrolled} and is modified to perform the mask prediction on the unrolled prediction, which simulates the masked input samples encountered at the intermediate refinement steps. On the other hand, the latter one learns to reconstruct the masked tokens in accordance with a selected context between two VL contexts that are mixed as a noisy input. This selective prediction improves cross-modal attention for an accurate generation. 

Through experiments on various downstream VL generation tasks, we empirically demonstrate that our MAGVLT significantly outperforms ARGVLT even with greatly reduced inference time. Especially, to the best of our knowledge, MAGVLT is the first model that obtains strong performances on both zero-shot I2T and zero-shot T2I generation tasks of MS-COCO benchmark \cite{coco} by a single moderate-sized model (fewer than 500M parameters) without relying on monomodal data and networks. Previously, as unified generative VL models, L-Verse \cite{kim2022verse} and UPGen \cite{anonymous2023connecting} have not showed zero-shot I2T results while OFA \cite{ofa} has used monomodal data and also has not showed zero-shot I2T and T2I results. Extensive ablations also validate the contribution of each component for MAGVLT.

To summarize, our main contributions are: (1) a masked generative VL transformer as a unified generative VL model that can produce both images and texts; (2) a robust training on image-text pairs by multiple training tasks that include the cross-modal mask prediction in tandem with the step-unrolled mask prediction and the selective prediction on the mixed context; and (3) an empirical validation of MAGVLT that outperforms the autoregressive model and moreover shows competitive performances on both of zero-shot I2T and T2I generation tasks for the first time without employing extra monomodal data and networks.


\section{Related Work}
\label{sec:background}

\subsection{Multimodal Vision-and-Language Modeling}

There has been a lot of research on multimodal VL training. Especially, in recent years, large-scale image-text datasets have made great progress in VL pretraining with various training objectives. For example, image-text matching \cite{lxmert, vilbert, e2e-vlp} and contrastive learning on image-text data \cite{coca, align, clip, ufo, Florence} has been widely used for discriminative representation learning. 

Since BERT \cite{bert} has shown impressive performances on many natural language processing tasks, masked language modeling has been widely adopted for VL pretraining. In particular, \cite{vl-t5} formulated multiple VL tasks as a text generation task and applied masked token prediction objective. \cite{vl-beit, beit3, m3ae, MultiMAE, maskvlm} have proposed to use a unified masked data modeling with a shared multimodal transformer while several algorithms have combined a number of objectives including image-text matching, contrastive VL loss, and masked language modeling \cite{dai2022enabling, vilt, vlmo, VinVL}. However, most of these masked VL pretraining algorithms have been developed for VL understanding tasks.

Meanwhile, AR generative modeling has also recently received lots of interest for VL pretraining due to its powerful generalization ability. While many algorithms have proposed to utilize the generative language modeling paradigm for VL understanding tasks \cite{git, hao2022language, wang2021simvlm, jin2021good, alayrac2022flamingo}, recent large-scale AR transformers trained on a large amount of image-text pairs have shown powerful performances for text-guided image generation \cite{dalle, parti, ding2021cogview, ding2022cogview2}. A few generative models based on AR decoding have been shown to produce both images and text sequences \cite{kim2022verse, ofa}, however, they have not shown competitive performances on both modalities.

\subsection{Non-Autoregressive Generative Modeling}
\begin{figure}[t]
\begin{minipage}{.47 \textwidth}\centering
    \includegraphics[width=1\textwidth]{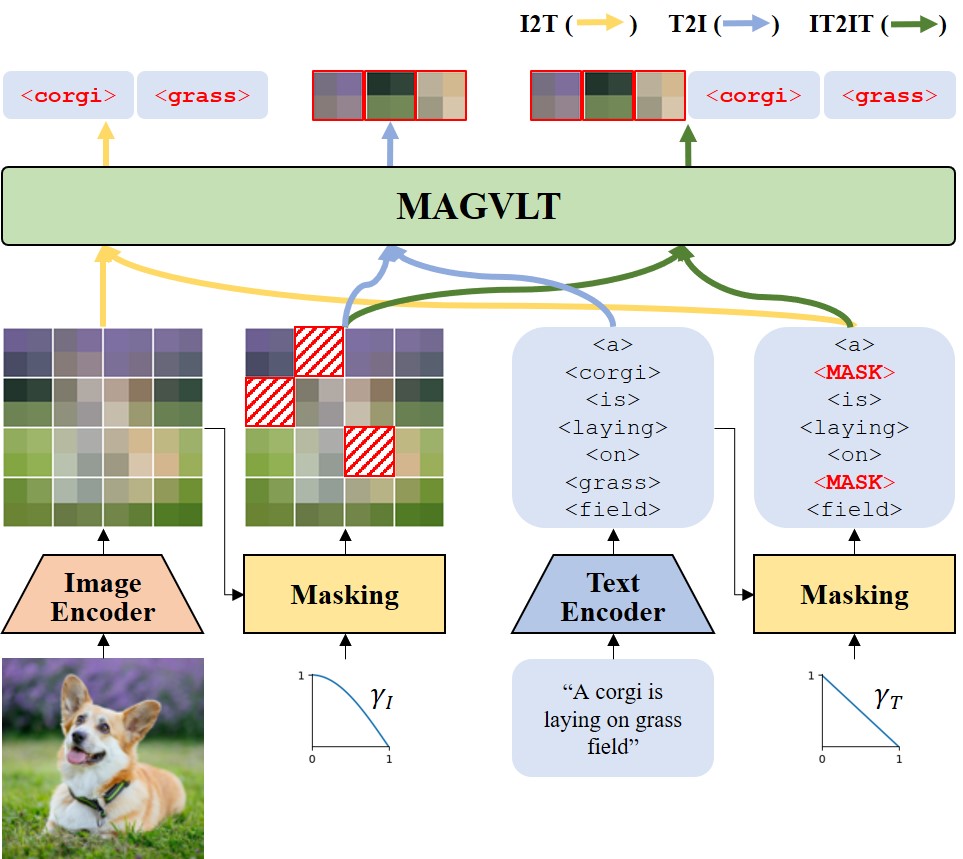}
    \caption{\label{fig:framework} Masked generative VL training via three multimodal masked token prediction tasks (I2T, T2I, IT2IT). Here, we represent VQ-GAN and BPE as image encoder and text encoder. MAGVLT predicts only masked tokens according to each task.}
\end{minipage}
\vspace{-0.4cm}
\end{figure}

Non-AR generative models have been increasingly used to lift certain limitations of AR models such as unidirectional attention and slow decoding. Among them, diffusion-based models have recently shown remarkable performances on the task of text-guided image generation \cite{nichol2021glide, dalle2, imagen, ERNIE-ViLG2, ldm}. However, for text generation, diffusion models \cite{Hoogeboom2021, d3pm} are still limited in achieving competitive performances compared to AR models.

Similar to diffusion-based generative models, masked generative models also perform iterative refinements for data generation and simulate various denoising steps during training. On top of that, masked generative models often conduct fewer refinement steps leading to faster generations. Therefore, masked generative models have recently been employed for language \cite{ghazvininejad-etal-2019-mask}, image \cite{chang2022maskgit, Draft-and-Revise}, and video processing \cite{maskvit, phenaki}. However, there have been almost no masked generative models that can generate both text and image data. Very recently, a concurrent work \cite{anonymous2023connecting} has tried to combine representation and generative learning for VL tasks into a single model that is based on masked token prediction. However, their generation performances are very poor on both modalities in contrast to the strong performances of MAGVLT on both modalities, and furthermore, our MAGVLT differs in that we use multiple cross-modal tasks for robust generative training on image-text pairs.

\section{Masked Generative Vision-and-Language Transformer} \label{sec:method}
\subsection{Masked Image-Text Modeling}

MAGVLT is based on the previous masked generative modeling for image and language processing \cite{chang2022maskgit, ghazvininejad-etal-2019-mask}. Given an image-text pair $(I,T)$, the image input $I$ is mapped to latent tokens $X=[x_i]_{i=1}^{N_I}$ by VQ-GAN \cite{Esser2021TamingTF}, where $N_I$ is the number of image tokens (\eg., $16 \times 16$), and the text sequence $T$ is also converted to the tokenized sequence $Y=[y_j]_{j=1}^{N_T}$ by byte pair encoding (BPE) \cite{bpe}, where $N_T$ is the number of text tokens (\eg, $64$). Then, $(X,Y)$ is fed into a bidirectional transformer with full attention. In contrast to the causal attention in the AR transformer, this full attention allows to fully utilize the whole context information for decoding and therefore leads to better output predictions. 
Here, to indicate which modality is processed by the shared transformer, we prepend learnable special tokens representing the modality such as {\fontfamily{qcr}\selectfont <BOI>} (\textit{begin-of-image}) and {\fontfamily{qcr}\selectfont <BOT>} (\textit{begin-of-text}) to the input tokens of each modality.


\begin{figure}[t]
\begin{minipage}{.47 \textwidth}\centering
    \includegraphics[width=1\textwidth]{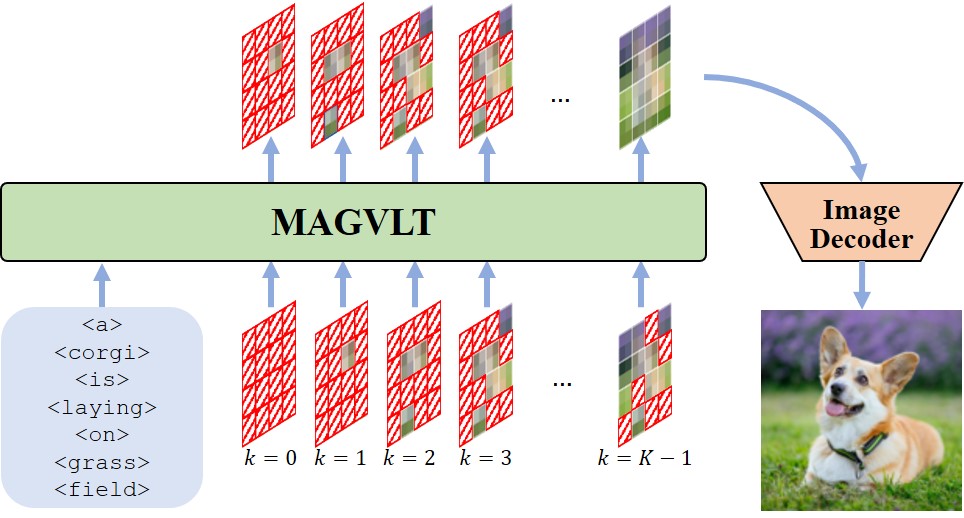}
    \caption{\label{fig:sampling} Iterative decoding process of MAGVLT for T2I task.}
\end{minipage}
\vspace{-0.3cm}
\end{figure}
For training MAGVLT on image-text pairs, we first sample a mask ratio $\gamma(r) \in(0,1]$ where $r \in[0,1)$ indicates the simulated refinement step ratio and is also uniformly sampled considering various steps during generation. Then, we uniformly sample $\lceil \gamma \cdot N \rceil$ tokens where $N=N_{I}+N_{T}$ and replace them with a special token {\fontfamily{qcr}\selectfont <MASK>}. Here, we separately apply this masking for each modality such that $\lceil \gamma_{I} \cdot N_{I} \rceil$ image tokens and $\lceil \gamma_{T} \cdot N_{T} \rceil$ text tokens are masked. Let $M_I=[m^I_i]_{i=1}^{N_I}$ and $M_T=[m^T_j]_{j=1}^{N_T}$ be the resulting binary image mask and binary text mask, respectively, such that $x_i$ is replaced with {\fontfamily{qcr}\selectfont <MASK>} if $m^I_i = 1$ while $y_j$ is replaced with {\fontfamily{qcr}\selectfont <MASK>} if $m^T_j = 1$. Note that we set $\gamma_{I}(\cdot)$ and $\gamma_{T}(\cdot)$ as the cosine function and the linear function, respectively, following \cite{chang2022maskgit} and \cite{ghazvininejad-etal-2019-mask}.

As shown in \cref{fig:framework}, MAGVLT is trained basically by the composition of three mask prediction losses: ${\mathcal L_{\text{I2T}}}$ for the I2T task, ${\mathcal L_{\text{T2I}}}$ for the T2I task, and ${\mathcal L_{\text{IT2IT}}}$ for the IT2IT task. And these losses are defined by the negative log-likelihood of the masked tokens:
\begin{eqnarray}
\label{eq:basic_loss}
{\mathcal L_{\text{I2T}}}\!\!&\!\!=\!\!&\!\!\! - \!\!\!\!\mathop{\mathbb{E}}_{(X,Y)\in {\mathcal D}} \bigg [ \sum_{\forall j \in [1, N_T], m^T_j=1} \!\!\!\!
\log p(y_j | Y_{{\bar M_T}}, X)
\bigg ], \\
{\mathcal L_{\text{T2I}}} \!\!&\!\!=\!\!&\!\!\! - \!\!\!\!\mathop{\mathbb{E}}_{(X,Y)\in {\mathcal D}} \bigg [ 
\sum_{\forall i \in [1, N_I], m^I_i=1} \!
\log p(x_i | X_{{\bar M_I}}, Y)
\bigg ], \\
{\mathcal L_{\text{IT2IT}}} \!\!&\!\!=\!\!&\!\!\! - \!\!\!\!\mathop{\mathbb{E}}_{(X,Y)\in {\mathcal D}} \bigg [
\sum_{\forall j \in [1, N_T], m^T_j=1} \!
\log p(y_j | Y_{{\bar M_T}}, X_{{\bar M_I}})\nonumber\\
&&~~~~~~~+\!\!\sum_{\forall i \in [1, N_I], m^I_i=1}
\log p(x_i | X_{{\bar M_I}}, Y_{{\bar M_T}})
\bigg ],
\end{eqnarray}
where ${\mathcal D}$ is the training dataset, and $X_{{\bar M_I}}$ and $Y_{{\bar M_T}}$ denote the masked image and the masked text, respectively.
\begin{figure}[t]
\begin{minipage}{.47 \textwidth}\centering
    \includegraphics[width=1 \textwidth]{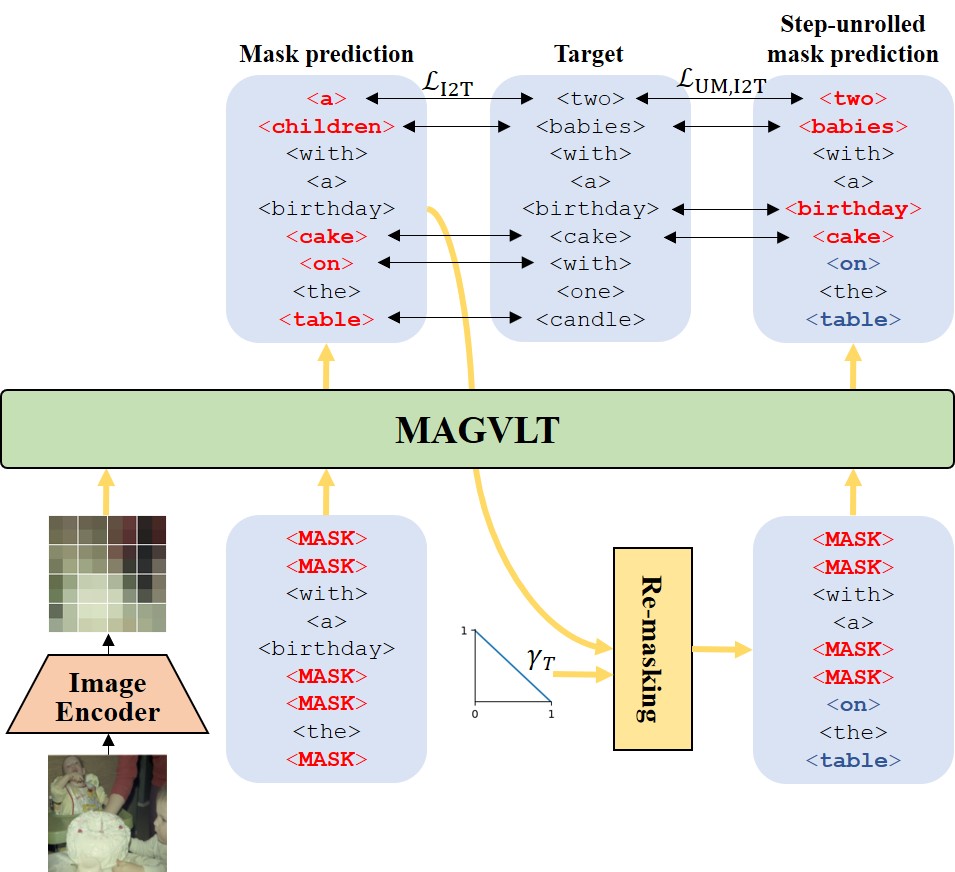}
    \caption{\label{fig:unroll} Step-unrolled mask prediction for I2T task. The one-step unrolled sequence is re-masked and then forwarded to the model. Throughout this process, the model faces inputs it would encounter during iterative inference.}
\end{minipage}
\vspace{-0.4cm}
\end{figure}

\subsection{Iterative Inference}

During inference, the target sequence is predicted by iterative decoding. The mask ratio is defined as a function of the decoding steps as $\gamma(\frac{k}{K})$ where $k \in \{0, 1, ..., K-1\}$ and $K$ is the total number of iterations.
For the first iteration ($k=0$), all the tokens are masked, and the model predicts all the tokens in parallel. 
For the next $k$th iteration, the most $\lceil \gamma(\frac{k}{K})N \rceil$ unconfident tokens are masked out and predicted again. 
This process is described in \cref{fig:sampling}. 
Here, it is noted that following the masking strategy in \cite{chang2022maskgit}, unmasked image tokens in previous steps are excluded in computing confidences and accordingly will never be masked again. On the other hand, unmasked text tokens can be selected as masked tokens again in the following iterations. Since the number of refinement steps is generally small (\eg, ~10), this iterative decoding with parallelizable predictions is significantly faster than the autoregressive decoding, especially when the number of tokens is very large.
\\[3pt]
\noindent\textbf{Target Length Prediction}. 
Since the length of text sequence is varied in contrast to the fixed number of image tokens, non-AR models like MAGVLT need to perform target length prediction for text generation. We follow the length predictor proposed in \cite{ghazvininejad-etal-2019-mask} where the output of {\fontfamily{qcr}\selectfont <BOT>} that is located between $X$ and $Y$ is the predicted text length. At test time, the text sequence is generated after the length is predicted. The loss of the auxiliary target length predictor, ${\mathcal L}_{\text{TL}}(N_T, {\hat N}_T)$, is defined by the cross entropy loss between the ground-truth length $N_T$ and the predicted length ${\hat N}_T$ given the maximum possible number of text tokens.
\begin{figure}[t]
\begin{minipage}{.47 \textwidth}\centering
    \includegraphics[width=.7\textwidth]{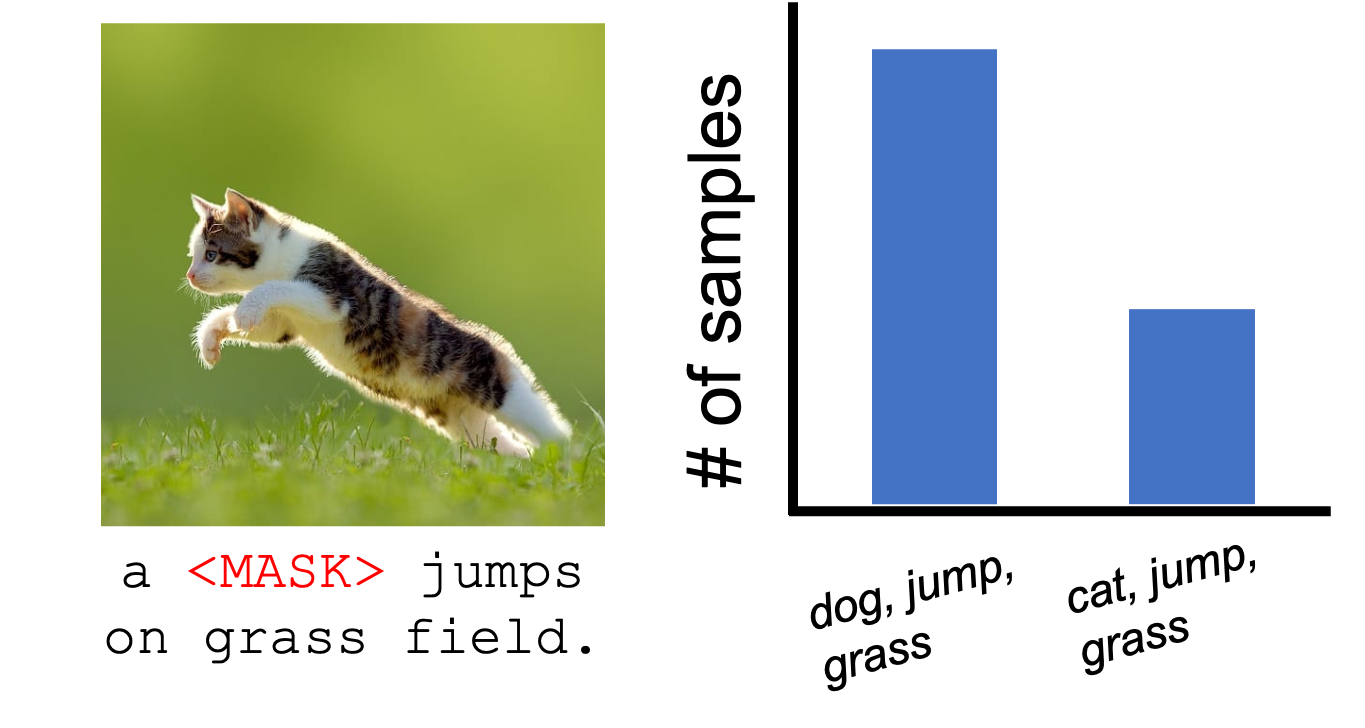}
    \caption{\label{fig:bias_stat}  An example of a paired data with masked caption \textbf{(Left)}, and biased statistics of word compositions  \textbf{(Right)}.}
\end{minipage}
\vspace{-0.4cm}
\end{figure}
\begin{figure*}[t]
\begin{minipage}{1 \textwidth}\centering
    \includegraphics[width=.95\textwidth]{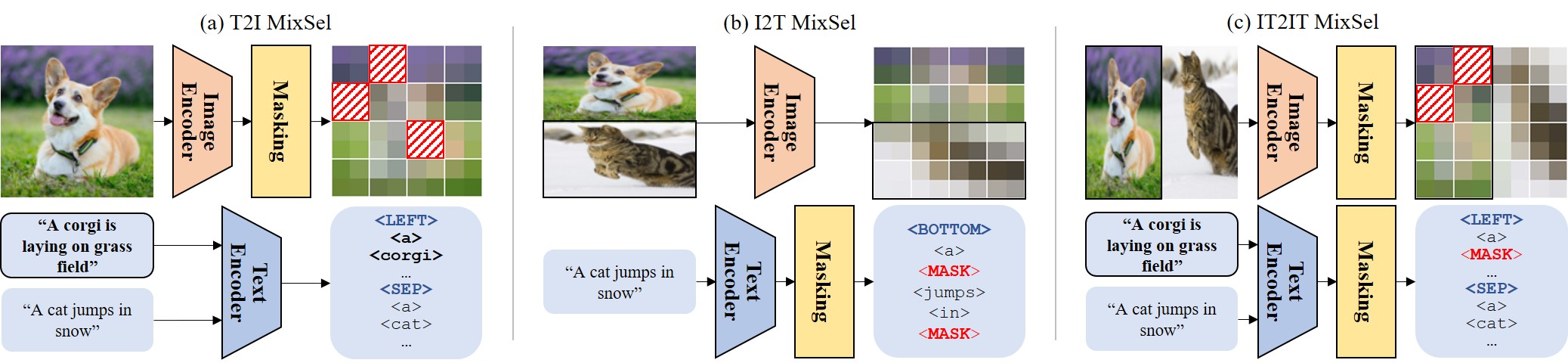}
    \caption{\label{fig:MixSel} \textit{MixSel} learning tasks corresponding to three multimodal tasks.}
\end{minipage}
\vspace{-0.4cm}
\end{figure*}
\subsection{Step-Unrolled Mask Prediction} \label{sec:unroll}
Although a variable mask ratio during training reflects various intermediate refinement steps, there still exists a gap between a corruption on the target tokens at training time and a corruption on the partially predicted tokens at test time.
SUNDAE \cite{savinov2022stepunrolled} tries to resolve this issue by optimizing the model conditioned on a corrupted target sequence which is sampled through one step generative unrolling during training and achieves significant performance improvements of the non-AR autoencoder for text generation.

Here, we adopt and modify this step-unrolled denoising as an additional training task for MAGVLT.
Since MAGVLT is based on the masked token prediction, we re-mask the one-step predicted sequence where the mask ratio is reduced from the previous mask ratio and then predict the re-masked tokens by MAGVLT. We call this task as step-unrolled mask prediction, dubbed UnrollMask. We apply UnrollMask only to the I2T and T2I training tasks to maintain the uncorrupted cross-modal context. \cref{fig:unroll} visually depicts this UnrollMask especially for the I2T task. We denote the UnrollMask loss as ${\mathcal L_{\text{UM}}}$, and for example ${\mathcal L_{\text{UM}}}$ on the I2T task can be defined as
\begin{eqnarray}
\label{eq:unroll_loss}
\!\!\!\!\!{\mathcal L_{\text{UM,I2T}}} \!\!&\!\!=\!\!&\!\!\! - \!\!\!\!\!\!\!\mathop{\mathbb{E}}_{(X,Y)\in {\mathcal D}} \!\! \bigg [
\sum_{\forall j \in [1, N_T ], {m^{T}_j}^{(+1)}=1} \!\!\!\!\!\!\!\!\!\!\!\!\!\!\!
\log p(y_j | {\hat Y}^{(+1)}_{{\bar M_T}^{(+1)}}, X)
\bigg ],
\end{eqnarray}
where ${\hat Y}^{(+1)}_{{\bar M_T}^{(+1)}}$ indicates the re-masked one-step unrolled prediction of $Y_{\bar M_T}$.

\subsection{Selective Prediction on Mixed Context} \label{sec:mixsel}
In this multimodal generative modeling, the model often ignores the cross-modal context and produces an output that is biased to the within-modal statistics. For example, in the I2T task of \cref{fig:bias_stat}, the model should predict the masked word token as `\textit{cat}' by the given image, however, the model often rather outputs `\textit{dog}' since `\textit{dog jumps}' is more likely occurred than `\textit{cat jumps}' before the text of `\textit{on grass field}' in the set of training text sequences.

Thus, in order to reduce such bias, we propose a simple yet effective additional learning task, named selective prediction on the mixed context (MixSel), which is described in \cref{fig:MixSel}.
As shown in the figure, two different input contexts are mixed in a half-and-half concatenated manner, and one of them is randomly selected to be the target context in generation. Here, a special token is appended to inform the selected context, for instance {\fontfamily{qcr}\selectfont <LEFT>} or {\fontfamily{qcr}\selectfont <RIGHT>} is used for the horizontally combined image or the concatenated text sequence while {\fontfamily{qcr}\selectfont <TOP>} or {\fontfamily{qcr}\selectfont <BOTTOM>} is used for the vertically combined image. Also, when two different text sequences are concatenated, another special token, {\fontfamily{qcr}\selectfont <SEP>}, is inserted between them.
The MixSel objective is denoted as ${\mathcal L_{\text{MS}}}$, and for instance ${\mathcal L_{\text{MS}}}$ on the I2T task can then be defined as
\begin{align}
\label{eq:hnh_loss}
&{\mathcal L_{\text{MS, I2T}}} =\nonumber\\~~ 
&- \!\!\!\!\mathop{\mathbb{E}}_{(X,Y)\in {\mathcal D}} \bigg [
\sum_{\forall j \in [1, N_T], m^T_j=1} \!
\log p(y^{\ell}_j | {\hat Y}^{\ell}_{{\bar M_T}}, \phi(X^1, X^2))
\bigg ],
\end{align}
where $\phi$ is the mixture function on the two images $X^1$ and $X^2$, and $\ell \in \{1,2\}$ represents the selected context. 


From this MixSel training task, the model is able to attend more carefully to the appropriate span of the cross-modal context and improve the accuracy of the cross-modal attention by mixing the original cross-modal content with randomly unrelated one. This could make the model to utilize the cross-modal context more trustfully
and hence more often in generation as the training progresses. Therefore, MixSel indeed helps in reducing the overlooking of the cross-modal context and circumventing the within-modal bias problem in test-time generation. Note that it is different from the previous mix-based data augmentation techniques \cite{mixup, cutmix, augmix} in that we retain the information of the original contexts entirely and randomly select the target one for generation. Moreover, although MixSel training is relevant to classifier-free guidance (CFG) \cite{ho_cfg} in that both try to strengthen the effect of the condition, MixSel does not have to perform the forward processing twice at
test time. Also, we can adapt CFG along with MixSel training.



\subsection{Multitask Pretraining}\label{sec:multitask}
As we mentioned above, MAGVLT is basically trained via three types of multimodal tasks: I2T, T2I, and IT2IT. 
During training, a task $\tau \in \{\text{I2T, T2I, IT2IT}\}$ is sampled from the categorical distribution with the predefined sampling probability $p_{\tau}$ for each iteration (batch-wise), and then apply the associated mask prediction loss ${\mathcal L_{\text{mask}, \tau}} \in \{{\mathcal L_{\text{I2T}}}, {\mathcal L_{\text{T2I}}}, {\mathcal L_{\text{IT2IT}}}\}$. Along with this mask prediction loss, we also add the target length prediction loss ${\mathcal L_{\text{TL}, \tau}}$, the UnrollMask loss ${\mathcal L_{\text{UM}, \tau}}$, and the MixSel loss ${\mathcal L_{\text{MS}, \tau}}$ according to the selected task $\tau$. Overall, the final objective of MAGVLT with respect to $\tau$ is
\begin{eqnarray} 
\label{eq:final_loss}
 {\mathcal L}_{\tau} &=& {\mathcal L}_{\text{mask}, \tau} + \lambda_{\text{TL}}{\mathcal L}_{\text{TL}, \tau} \cdot {\mathbb I} [\tau \neq \text{T2I}] \nonumber \\ &&+ \lambda_{\text{UM}}{\mathcal L}_{\text{UM}, \tau} \cdot {\mathbb I}[\tau \neq \text{IT2IT}] + \lambda_{\text{MS}} {\mathcal L}_{\text{MS}, \tau},
\end{eqnarray}
where ${\mathbb I}[\cdot]$ is the indicator function, and $\lambda_{\text{TL}}$, $\lambda_{\text{UM}}$, and $\lambda_{\text{MS}}$ are relative loss weights. Here, we fix $\lambda_{\text{TL}}=0.01$, $\lambda_{\text{UM}}=1.0$, and $\lambda_{\text{MS}}=0.5$ for all our experiments.

\section{Experiment}
\label{sec:exp}
In this section, we elaborate the experiments on the VL generation tasks with extensive ablation studies to identify the contribution of each factor in the proposed algorithm. Official codes will be available\footnote{\url{https://github.com/kakaobrain/magvlt}}.
\subsection{Experimental Setup}
\noindent\textbf{Model}.
VLTs (\ie ARGVLT and MAGVLT) have 447M parameters (24 layers, 1024 hidden dimension, and 8 attention heads) including VQ-GAN in total.
We also perform experiments about scaling of VLTs, and the results are presented in Appendix.
As an image encoder, VQ-GAN \cite{Esser2021TamingTF} converts a 256$\times$256 image into 16$\times$16 tokens with 16,384 codebook size. 
For text sequence, we adopt the BPE tokenizer \cite{bpe} used in CLIP \cite{clip} with 49,408 vocabulary size. We fix the text sequence length to 64.
\\[3pt]
\noindent\textbf{Dataset}. 
We pretrain ARGVLT and MAGVLT from scratch using paired image-text datasets. 
Our pretraining data consists of Conceptual Captions 3M (CC3M)\cite{cc3m}, Conceptual Captions 12M (CC12M)\cite{cc12m}, SBU Caption\cite{NIPS2011_SBU}, and Visual Genome\cite{krishna_VG} datasets.
Together, there are about 17M image-text pairs.
\\[3pt]
\noindent\textbf{Pretraining}. 
There are many options to train VLTs. 
Note that T2I and I2T are available for both ARGVLT and MAGVLT while IT2IT is only available for MAGVLT.
We experiment various subsets of the multimodal tasks in order to investigate the effectiveness of each task.
In specific, we pretrain ARGVLT on three different subsets which consist of  \textit{T2I only}, \textit{I2T only}, and \textit{T2I} \& \textit{I2T}. 
Likewise, we pretrain MAGVLT on three different subsets which consist of  \textit{T2I only}, \textit{I2T only}, and \textit{T2I} \& \textit{I2T} \& \textit{IT2IT}.
More details of pretraining will be found in Appendix.
\\[3pt]
\noindent\textbf{Evaluation}. 
We compare generative VL models based on cross-modality generation tasks especially in \textit{zero-shot} settings in order to evaluate the generalization ability of the proposed method.
We also provide more results including finetuning VLTs on downstream tasks in Appendix.
\\[3pt]
\noindent\textbf{Sampling}. 
For text-to-image generation, following\cite{kim2022verse},  we obtain 32 samples from each trained VLT (\ie ARGVLT and MAGVLT) and calculate similarity scores between the sampled images and the conditioning text by CLIP \cite{clip} to select a top ranked image (\textit{clip reranking}). 
Likewise, for image-to-text generation (image captioning), we produce 64 samples from each trained VLT and select a top ranked text by CLIP scores. The number of refinement steps is set to $K=10$ and $K=12$ for image generation tasks and text generation tasks, respectively.

\begin{figure}[t]
\begin{minipage}{.45 \textwidth}\centering
    \includegraphics[width=.95\textwidth]{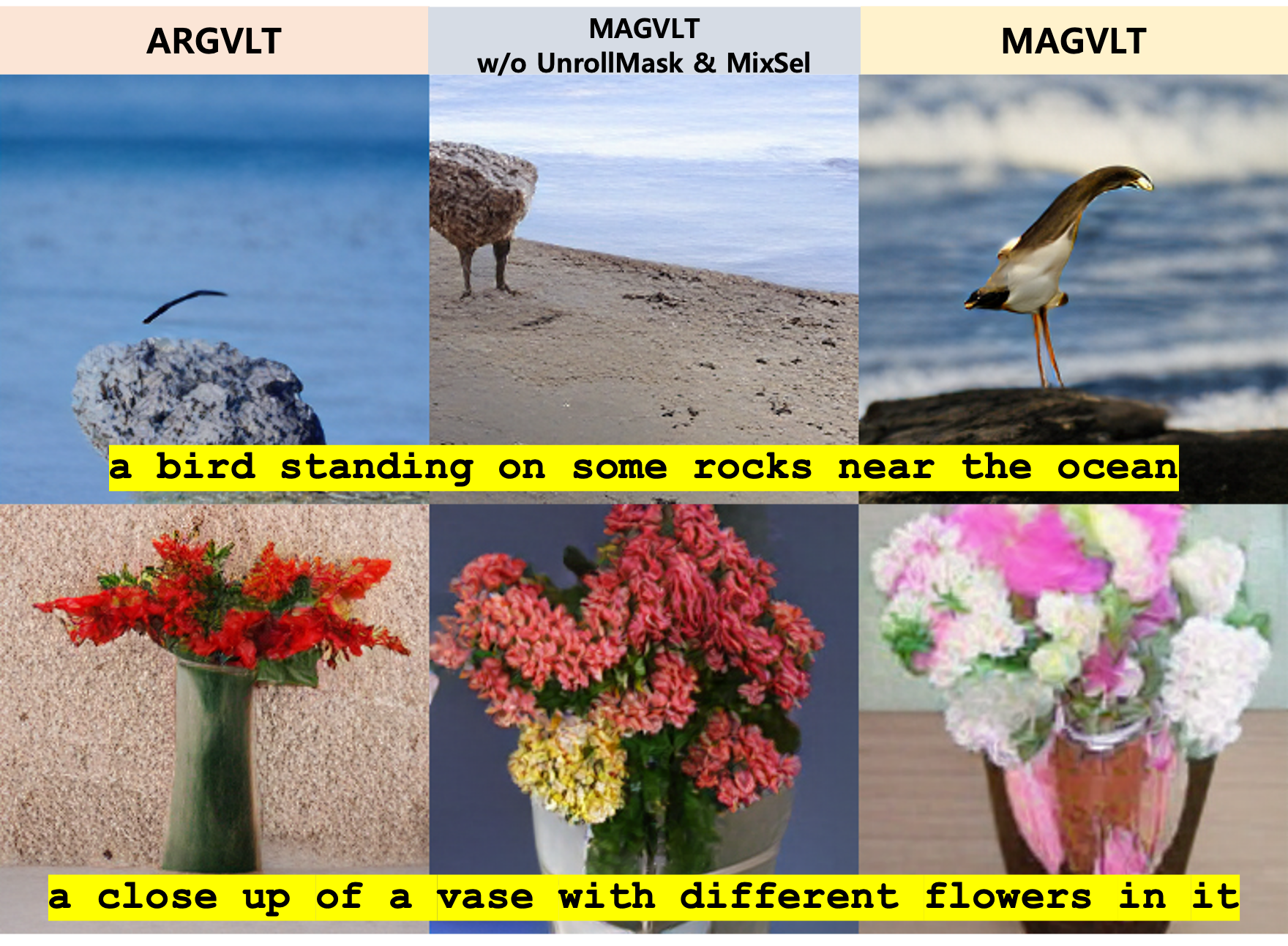}
    \caption{\label{fig:t2i} Text-to-Image samples on MS-COCO captions. The images in the second column are sampled from MAGVLT trained without \textit{UnrollMask} and \textit{MixSel}.
    MAGVLT generated more appropriate images on the corresponding caption.
    More samples will be found in Appendix.
    }
\end{minipage}
\vspace{-0.4cm}
\end{figure}


\bgroup
\def\arraystretch{1.05}%
\begin{table}[t]
\centering
\small
\begin{tabular}{lccc}
\hline
Model & FID ($\downarrow$) & IS ($\uparrow$) & Speed\\
\hline
\rowcolor[gray]{0.85}\multicolumn{4}{l}{\textit{\textbf{AR based}}} \\
CM3-Medium (2.7B) \cite{aghajanyan2022cm3} & 36.78  & -  & - \\
DALL-E (12B) \cite{dalle} & 27.5  & 17.9 & - \\
CogView (4B) \cite{ding2021cogview} & 27.1  & 18.2  & -\\
CogView2 (6B) \cite{ding2022cogview2} & 24.0  & 22.4  & - \\
Parti-350M (350M) \cite{parti} & 14.10  & -  & - \\
Make-A-Scene (4B) \cite{gafni2022make} & \bf{11.84}  & -  & - \\
ARGVLT (\textit{T2I only}) (447M)  & 21.80 & 19.27 & $1.00\times$ \\
\hline
\rowcolor[gray]{0.85}\multicolumn{4}{l}{\textit{\textbf{Non-AR based}}} \\ 
GLIDE (3.5B) \cite{nichol2021glide} & 12.24  & - & -  \\
DALL-E-2 (6.5B) \cite{dalle2} & 10.39  & - & - \\
Imagen (4.9B) \cite{imagen} & 7.27  & - & - \\
ERNIE-ViLG 2.0 (24B) \cite{ERNIE-ViLG2} & {\bf 6.75}  & - & - \\
{\bf MAGVLT (\textit{T2I only}) (447M)}  & 10.74 & 23.94 & $8.12\times$ \\
\hline
\rowcolor[gray]{0.85}\multicolumn{4}{l}{\textit{\textbf{Available for both T2I \& I2T}}} \\ 
UPGen (307M) \cite{anonymous2023connecting} & 65.25  & -  & - \\
L-Verse (500M) \cite{kim2022verse} & 37.2  & -  & - \\
ARGVLT (447M) & 16.93 & 22.50 & $1.00\times$ \\
{\bf MAGVLT (447M)}  & \bf{12.08} & \bf{22.75} & $8.12\times$ \\
\hline
\end{tabular}
\caption{\textit{Zero-shot} T2I results on MS-COCO validation set. Here, we compute FID and IS on a subset of 30,000 captions sampled from COCO validation. \label{tab:t2i}}
\end{table}
\egroup
\bgroup
\def\arraystretch{1.05}%
    \begin{table}[t]
    \centering
    \small
    \begin{tabular}{lcc}
    \hline
    Model & FID ($\downarrow$) & IS ($\uparrow$) \\
    \hline
    ARGVLT & 5.62   & 29.21 \\
    {\bf MAGVLT}  &  {\bf 3.17} & {\bf 30.79} \\
    \hline
    \end{tabular}
    \caption{\textit{Zero-shot} image inpainting results on MS-COCO validation. \label{tab:inpainting}}
    \vspace{-0.6cm}
    \end{table}
\egroup

\subsection{Image Generation}
\label{sec:image gen}
\noindent\textbf{Text-to-Image}. 
We evaluate the zero-shot generalization capability of MAGVLT under text-to-image generation.
We measure quantitative metrics of quality of generated images by Fréchet Inception Distance (FID) \cite{FID} and Inception Score (IS) \cite{salimans2016improved} on MS-COCO \cite{coco} validation.
In addition, we compare the relative decoding speed of MAGVLT against ARGVLT. The results are shown in \cref{tab:t2i}, where the models are grouped according to: (1) whether they are AR-based or not, and (2) the modality they can generate; the models in the first two groups are able to generate image only while the models in the last group can generate both image and text.
MAGVLTs significantly outperform AR-based methods including ARGVLTs as well as obtain comparable scores to the state-of-the-art diffusion-based methods. 
Note that all the other models in the non-AR group have more than 1B parameters while MAGVLTs have less than 500M parameters. And the performance gap between the task-specific MAGVLT (T2I only) and the universal MAGVLT is small.
Moreover, MAGVLTs generate an image more than eight times faster than ARGVLTs.
We provide qualitative samples in \cref{fig:t2i}.
\\[4pt]
\noindent\textbf{Image Inpainting}. 
One of the key advantages of MAGVLT over ARGVLT is that it enables bidirectional encoding of conditional information. MaskGIT \cite{chang2022maskgit} already demonstrated this advantage. To reconfirm it on MAGVLT, we conduct similar image inpainting experiments. In detail, the central 8$\times$8 image tokens corresponding to the central 50\% of the whole 16$\times$16 image tokens are masked out, and then replaced with newly-generated tokens conditioning on the unmasked image tokens and the ground-truth text tokens. The output images are blended with the input images along the mask boundary following \cite{chang2022maskgit}.
Quantitatively, as shown in \cref{tab:inpainting}, MAGVLT outperforms ARGVLT. which is also observed in qualitative samples of \cref{fig:t2i_inpaint_more} in Appendix.

\bgroup
\def\arraystretch{1.05}%
\begin{table}[t]
\centering
\small
\setlength{\tabcolsep}{3pt}
\scalebox{1}{
    \begin{tabular}{lccccc}
    \hline
    \multicolumn{1}{l}{Model} & B-4  & M    & C  & S & Speed
    \\ \hline
    \rowcolor[gray]{0.85}\multicolumn{6}{l}{\textit{\textbf{with external language model}}} \\ 
    ZeroCap (345M) \cite{tewel2022zerocap} &2.6 & 11.5 & 14.6 & 5.5 & - \\
    MAGIC (1.5B) 
    \cite{su2022magic} & 12.9& 17.4 &  49.3& 11.3 & -\\
    VLKD$_{\text{ViT-B/16}} $ (406M) 
    \cite{dai2022enabling} & {\bf 16.7} & {\bf 19.7} & 58.3 & {\bf 13.4} & -\\
    Flamingo-3B (3B) \cite{alayrac2022flamingo} &- & -& {\bf 73.0} & - & -\\
    \hline
    \rowcolor[gray]{0.85}\multicolumn{6}{l}{\textit{\textbf{without external language model}}} \\ 
    SimVLM$_{\text{huge}}$ (632M) \cite{wang2021simvlm} & 11.2& 14.7 & 32.2 & 8.5 & -\\
    ARGVLT (\textit{I2T only}) (447M) & 11.4 & 15.1 & 47.4 & 11.4 & $1.00\times$ \\
    ARGVLT (447M) & 10.9 & 14.9 & 45.5 & 11.2 & $1.00\times$ \\
    {\bf MAGVLT (\textit{I2T only})} (447M)& 12.9& 17.1& 53.5& 12.9 & $1.56\times$\\
    {\bf MAGVLT} (447M)& {\bf 14.6} & {\bf 19.0} & {\bf 60.4} & {\bf 14.3} & $1.56\times$
    \\ \hline
    \end{tabular}
}
\caption{\textit{Zero-shot} I2T results on MS-COCO Karpathy test. \label{tab:i2t_coco}}
\vspace{-0.1cm}
\end{table}
\egroup

\begin{figure}[t]
\begin{minipage}{.47 \textwidth}\centering
    \includegraphics[width=.95\textwidth]{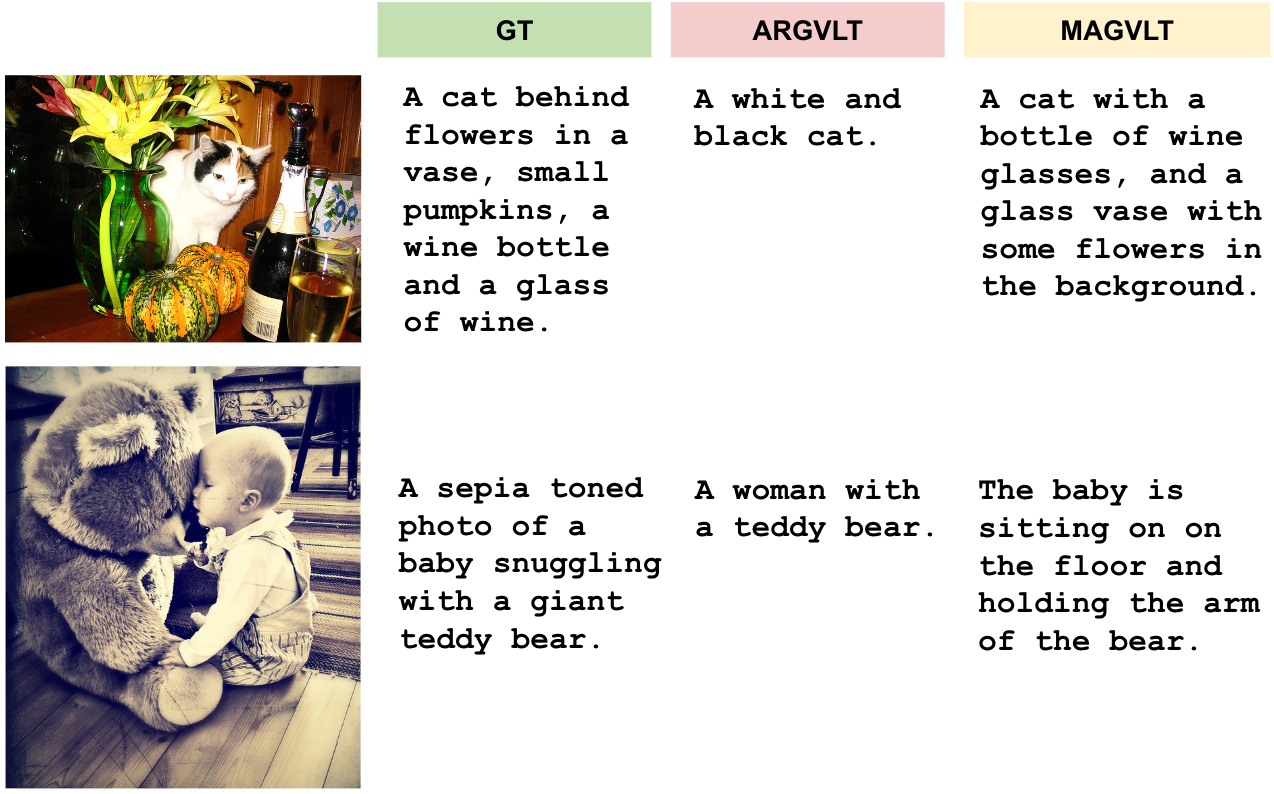}
    \caption{\label{fig:i2t} Image-to-text samples on MS-COCO images. MAGVLT generated more proper captions on the corresponding images. More samples will be found in Appendix.}
\end{minipage}
\vspace{-0.2cm}
\end{figure}


\subsection{Text Generation}
\label{sec:text gen}
\noindent\textbf{Image Captioning}. 
We evaluate the zero-shot image caption generation on MS-COCO Caption\cite{coco}, and NoCaps\cite{nocaps}. 
We measure quantitative metrics of the quality of the generated caption compared to the ground truth by BLEU-4 (B-4), METEOR (M), CIDEr (C), and SPICE (S).
The evaluation results on MS-COCO are shown in \cref{tab:i2t_coco}. 
Likewise as in text-to-image generation, we can verify that MAGVLTs improve performances over ARGVLTs significantly.
Moreover, MAGVLT outperforms some baselines that leverage external language models in CIDEr and SPICE which are specifically designed for the captioning task.
Note that MAGVLT has about six times smaller parameters than Flamingo-3B \cite{alayrac2022flamingo}, and even can generate images by a single model. In addition, the universal MAGVLT performs better than the task-specific MAGVLT (I2T only) maybe due to the synergetic improvement between the modalities. Regarding the marginal speedup ($1.56\times$) by MAGVLT over ARGVLT for I2T, compared to the fixed number of image tokens (256), the numbers of text tokens are quite small and generally less than 32, thus, the relative speedup by parallel predictions for
12 steps is reduced in text generation. Here, MAGVLT generates a text sequence given the pre-predicted target length. We provide qualitative samples in \cref{fig:i2t}.

Here, since L-Verse \cite{kim2022verse}, which can generate both modalities, provides I2T results only by scratch training on MS-COCO, for more comparison to L-Verse, we also perform the training from scratch on MS-COCO alone and obtain a much better FID score for T2I but a slightly lower CIDEr score for I2T compared to L-Verse (18.49 vs. 45.8 in FID, 85.3 vs. 102.2 in CIDEr).

\bgroup
\def\arraystretch{1.05}%
\begin{table}[t]
\centering
\small
\begin{tabular}{lcc}
\hline
Model & CIDEr & SPICE
\\ \hline
\rowcolor[gray]{0.85}\multicolumn{3}{l}{\textit{\textbf{with MS-COCO in training}}} \\ 
FewVLM$_{\text{large}}$ (740M)\cite{jin2021good} & 47.7& {\bf 9.1} \\
MetaLM (545M) 
\cite{hao2022language} 
& {\bf 58.7}& 8.6 \\
\hline
\rowcolor[gray]{0.85}\multicolumn{3}{l}{\textit{\textbf{without MS-COCO in training}}} \\ 
VLKD$_{\text{RN50x16}} $ (406M)\cite{dai2022enabling} & 54.0 & {\bf 9.6} \\
SimVLM$_{\text{huge}}$ (632M)\cite{wang2021simvlm} & {\bf 101.4} & -\\
ARGVLT (\textit{I2T only}) (447M) & 34.8& 6.5 \\
ARGVLT (447M) & 33.4& 6.4  \\
{\bf MAGVLT (\textit{I2T only})} (447M)& 37.7 & 7.2  \\
{\bf MAGVLT} (447M)& 46.3 & 8.7 
\\ \hline
\end{tabular}
\caption{\textit{Zero-shot} I2T results on NoCaps validation. \label{tab:i2t_nocaps}}
\end{table}
\vspace{-0.2cm}
\egroup
The evaluation results on NoCaps are shown in \cref{tab:i2t_nocaps}.
Basically, including MS-COCO in training is beneficial in performing on NoCaps since the interface of the caption collection for NoCaps closely resembles that used for the collection of the MS-COCO captions. 
Yet, MAGVLT shows comparable performances compared to FewVLM \cite{jin2021good}.
Also in this task, MAGVLT significantly outperforms ARGVLT.
MAGVLTs underperform in comparisons to VLKD \cite{dai2022enabling} and SimVLM \cite{wang2021simvlm}.
This might be due to that VLKD (RN50x16) has larger parameters ($\simeq$700M) than MAGVLT and also leverages an external language model (BART\cite{lewis-etal-2020-bart}) unlike MAGVLT.
SimVLM uses ALIGN dataset \cite{align} which contains 1.8B image-text pairs as well as C4 \cite{c4data} dataset which contains 360M text-only instances while MAGVLT uses only 17M image-text pairs for pretraining.
\\[5pt]
\noindent\textbf{Text Infilling}. 
Similar to image inpainting, we conduct the text infilling experiment where the central 50\% part of the text is erased by a mask and then replaced with generated tokens from the trained model. The qualitative samples are shown in \cref{fig:i2t_infill_more} in Appendix, where we can see that the infilled words generated by MAGVLT are better aligned with surrounding context words, compared to ARGVLT. Also, as shown in \cref{tab:infilling}, MAGVLT quantitatively outperforms ARGVLT. 

\bgroup
\def\arraystretch{1.05}%
    \begin{table}[t]
    \centering
    \small
    \begin{tabular}{lcccc}
    \hline
    \multicolumn{1}{l}{Model} & B-4  & M & C  & S\\
    \hline
    ARGVLT & 39.5  & 32.2 & 119.8 & 23.8\\
    {\bf MAGVLT} & {\bf 42.7}  & {\bf 35.2} & {\bf 135.3} & {\bf 26.3}\\
    \hline
    \end{tabular}
    \caption{\textit{Zero-shot} text infilling results on MS-COCO Karpathy test. \label{tab:infilling}}
    \vspace{-0.3cm}
    \end{table}
\egroup

\subsection{Ablation Studies} \label{sec:ablation}
\bgroup
\def\arraystretch{1.05}%
    \begin{table}[t]
    \centering
    \small
    \begin{tabular}{lccc}
    \hline
    \multicolumn{1}{c}{Task sample weights} & CIDEr ($\uparrow$)  & FID ($\downarrow$)\\
    {\fontfamily{qcr}\selectfont T2I:I2T:IT2IT} & &  \\
    \hline
    {\fontfamily{qcr}\selectfont 1:0:0} (\textit{T2I only}) & - & \bf{10.74}\\
    {\fontfamily{qcr}\selectfont 0:1:0} (\textit{I2T only}) & 53.5 & - \\
    {\fontfamily{qcr}\selectfont 0:0:1} (\textit{IT2IT only}) & 55.3 & 12.06\\
    {\fontfamily{qcr}\selectfont 8:2:0} (\textit{T2I} \& \textit{I2T}) & 59.7 & 13.09 \\
    {\fontfamily{qcr}\selectfont 2:1:1} & \bf{61.7} & 15.17 \\
    {\fontfamily{qcr}\selectfont 6:1:1} & 60.7 & 12.65 \\
    {\fontfamily{qcr}\selectfont 8:1:1}$^*$ & 60.4  & 12.08 \\
    {\fontfamily{qcr}\selectfont 10:1:1} & 59.2 & 12.07 \\
    \hline
    \end{tabular}
    \caption{Variants of MAGVLT. \label{tab:variants}}
    \vspace{-0.3cm}
    \end{table}
\egroup
\noindent\textbf{Variants of MAGVLT.}
Here, we investigate the effects of sampling weights for three multimodal tasks corresponding to $p_{\tau}$ in learning MAGVLT, and the results are shown in \cref{tab:variants}.
We denote the three weights in a form of {\fontfamily{qcr}\selectfont T2I:I2T:IT2IT} in the table.
The trained model by T2I only produces the best performance for text-to-image generation. 
However, in contrast to this, the trained model by I2T only shows the worst performance for captioning. 
We observe that the training loss of the mask prediction of this I2T only model is the lowest compared to the other models. 
This may suggest that the bias issue we mentioned in \cref{sec:mixsel} can be more serious, especially in non-AR methods.
As shown in the result, learning T2I along with I2T resolves the issue to some extent. 
The ratio of {\fontfamily{qcr}\selectfont 8:2:0} shows better I2T performance in CIDEr but inferior
T2I performance in FID than that of {\fontfamily{qcr}\selectfont 0:0:1}. We observe that the inclusion of the uncorrupted cross-modal context is necessary. The model trained with {\fontfamily{qcr}\selectfont 2:1:1} weights shows the best captioning performance but the worst T2I performance at a time.
As the portion of T2I training is getting larger, the model performs better in T2I but worsens in I2T. 
It means that there is a trade-off between T2I and I2T according to the sampling ratio.
We use the most balanced one ({\fontfamily{qcr}\selectfont 8:1:1}) super-scripted by * as the default setting for MAGVLT.

It is noted that regarding the performance drop by IT2IT modeling rather than T2I-only modeling for the T2I task, compared to I2T, in T2I the benefit of focusing on the cross-modal context is less significant, especially in the later refinement steps. Therefore, improved cross-modal attention by IT2IT training could be less effective for the T2I task. Having said that, the performance drop by MAGVLT for T2I is small even though it can also perform I2T. And we observe that in terms of CLIP scores, our IT2IT training is slightly better than T2I-only training (0.3176 vs. 0.3145) due to the enhanced cross-modal alignment. Moreover, the capacity of our moderate-sized model would be still limited in generating both modalities. Under this limited capacity, there exist some performance trade-offs, and here it would be more leaned to I2T. \cref{tab:scaling_t2i} in Appendix shows that when we increase the model capacity about two times, the large model with IT2IT performs better than the medium model with T2I-only training on the T2I task.
\bgroup
\def\arraystretch{1.05}%
    \begin{table}[t]
    \centering
    \small
    \begin{tabular}{lcc}
    \hline
    \multicolumn{1}{l}{Model} & CIDEr ($\uparrow$)  & FID ($\downarrow$) \\
    \hline
    MAGVLT (\textit{T2I only}) & - & \bf{10.74}\\
    ~~w/o MixSel & - & 10.97 \\
    ~~w/o UnrollMask and MixSel & - & 11.72 \\
    MAGVLT (\textit{I2T only}) & \bf{53.5} & - \\
    ~~w/o MixSel & 51.3 & - \\
    ~~w/o UnrollMask and MixSel & 48.0 & - \\
    MAGVLT & \bf{60.4} & \bf{12.08} \\
    ~~w/o MixSel & 58.9 & \bf{12.07} \\
    ~~w/o UnrollMask & 56.5 & 13.26 \\
    ~~w/o UnrollMask and MixSel & 53.8 & 14.12 \\
    \hline
    \end{tabular}
    \caption{Effectiveness of additional training tasks. \label{tab:addtask}}
    \vspace{-0.3cm}
    \end{table}
\egroup
\\[5pt]
\noindent\textbf{Effectiveness of Additional Tasks}. 
Here, we investigate the effects of the additional tasks (\ie \textit{UnrollMask} and \textit{MixSel}), and the results are 
 shown in \cref{tab:addtask}.
MAGVLTs without the additional tasks are clearly underperformed in total.
Applying UnrollMask in pretraining significantly improves the performances of the base models on both T2I and I2T tasks.
Including MixSel along with UnrollMask also improves the performances of the T2I-only model and especially the models on I2T. Similar to IT2IT training, this improved cross-modal
attention by MixSel could be less effective for T2I task compared to I2T task. We also experimentally found that the performance gain by MixSel is somewhat marginal for ARGVLT.
We hypothesize that the causal attention of AR models would be hard to encode the randomly changing span of context.

\section{Conclusion} \label{sec:concls}

In this work, we propose MAGVLT as a unified generative VL model that can produce both image and text data. MAGVLT is robustly trained on image-text pairs by multiple cross-modal tasks and significantly outperforms ARGVLT achieving strong performances on both of zero-shot I2T and T2I tasks. In future work, we first need to scale-up MAGVLT in terms of both model and data for better generalizability. Also, we aim to leverage a pretrained language model to enhance natural language processing and eventually to enable in-context VL learning. We will also try to exploit an encoder-decoder architecture to robustly perform on both understanding and generation tasks.

{\small
\subsection*{Acknowledgements}
We would like to thank Brain Cloud Team at Kakao Brain for their support. This work was also supported by Korea University Grant (K2304351) and Institute of Information \& communications Technology Planning \& Evaluation (IITP) grant funded by the Korea government (MSIT) (No. 2022-0-00612, Geometric and Physical Commonsense Reasoning based Behavior Intelligence for Embodied AI).
}

{\small
\bibliographystyle{ieee_fullname}
\bibliography{egbib}
}

\newpage
\appendix

\section{Training Details} 
We apply BPE dropout with a rate of 0.1. We also apply residual and attention dropouts with a rate of 0.1, and label smoothing for both image and text loss computation with a rate of 0.1. We train both ARGVLT and MAGVLT models using AdamW optimizer with $\beta_1 = 0.9$, $\beta_2 = 0.96$, $\epsilon = 10^{-8}$, weight decay coefficient of $4.5\times 10^{-2}$, and the learning rate of $4.5\times 10^{-4}$ with a cosine annealing. The gradients are clipped by a norm using a threshold of 4, prior to applying the Adam update. When training ARGVLT, we observe that calculating the predictive losses on the context tokens along with the generation tokens improves the overall performance. Hence, we compute the losses on the whole concatenated token sequence with the loss 
coefficients to 0.9 and 0.1 for generation modality and conditional modality, respectively. The data augmentations used in \cite{dalle} are applied to the images before encoding them using VQ-GAN. 
For positional embedding, we adopt a learnable absolute position encoding, for both image and text modalities. The encoded image tokens are flattened by the raster scan order before being fed into the transformer. MAGVLT was trained on 128 V100 GPUs for 40K updates with a batch size of 4,096, which takes about 3 days.

\cref{tab:arch_hparam} describes the detailed architecture hyperparameters for the transformers we used including the large models. 
 
\bgroup
\def\arraystretch{1.05}%
    \begin{table}[h]
    \centering
    \small
    \begin{tabular}{ccc}
    \hline
    \bf{Parameter} & \multicolumn{2}{c}{\textbf{Model}} \\
    \cmidrule(r){2-3}
    & ARG/MAGVLT & ARG/MAGVLT$_{\text{Large}}$ \\
    \hline
    Params & 371M & 840M \\  
    Layers & 24 & 36  \\ 
    Embed Dim & 1024 & 1280 \\
    Heads & 8 & 10 \\ 
    \hline
    \end{tabular}
    \caption{Detailed architecture hyperparmeters. The left model column represents the default model described in the main paper, while the right column indicates the large model that will be presented in the next section.
    \label{tab:arch_hparam}}
    \end{table}
\egroup

\section{Model Scaling}
It is well known that scaling up the pretrained generative model generally improves the generalization ability, and recently VL models often have more than 1B parameters. Therefore, we also scale up our VLTs and evaluate those for the tasks of zero-shot I2T and T2I on MS-COCO. As shown in \cref{tab:arch_hparam}, the large model (MAGVLT$_\text{Large}$) contains 840M parameters for the transformer and 916M parameters including VQ-GAN in total. 
MAGVLT$_\text{Large}$ was trained on 128 V100 GPUs for 80K updates with a batch size of 4096, which takes about 12 days.
\bgroup
\def\arraystretch{1.05}%
    \begin{table}[h]
    \centering
    \small
    \begin{tabular}{lccc}
    \hline
    Model & FID ($\downarrow$) & IS ($\uparrow$) & Speed\\
    \hline
    ARGVLT  & 16.93 & 22.50 & 1.00$\times$ \\
    ARGVLT$_\text{Large}$  & 13.01 & 23.75 & 0.51$\times$ \\ 
    MAGVLT  & 12.08 & 22.75 & \bf{8.12}$\times$ \\
    MAGVLT$_\text{Large}$  & \bf{10.14} & \bf{25.15} & 6.97$\times$ \\
    \hline
    \end{tabular}
    \caption{\textit{Zero-shot} T2I results on MS-COCO validation. \label{tab:scaling_t2i}}
    \end{table}
\egroup

The zero-shot T2I results on MS-COCO are shown in \cref{tab:scaling_t2i}. Notably, the large-scale models of both VLTs significantly improve FID and IS scores with large margin, compared to their respective default models. In addition, the degree of sampling speed reduction by model scaling is relatively smaller in MAGVLT than that in ARGVLT. Note that  MAGVLT$_\text{Large}$ is slightly slower than the default MAGVLT (6.97$\times$ vs 8.12$\times$), however it is still much faster than the default ARGVLT which has much fewer parameters.  
\bgroup
\def\arraystretch{1.05}%
\begin{table}[h]
\centering
\small
\scalebox{1}{
    \begin{tabular}{lcc}
    \hline
    \multicolumn{1}{l}{Model} & CIDEr  & SPICE
    \\ \hline
    \rowcolor[gray]{0.85}\multicolumn{3}{l}{\textit{\textbf{MS-COCO}}} \\ 
    ARGVLT & 45.5 & 11.2 \\
    ARGVLT$_\text{Large}$ & 43.6 & 11.2 \\ 
    MAGVLT & 60.4 & 14.3 \\
    MAGVLT$_\text{Large}$ & \bf{68.1} & \bf{15.5} \\ 
    \hline
    \rowcolor[gray]{0.85}\multicolumn{3}{l}{\textit{\textbf{NoCaps}}} \\ 
    ARGVLT & 33.4 & 6.4  \\
    ARGVLT$_\text{Large}$ & 34.1  &  6.1 \\ 
    MAGVLT & 46.3 & 8.7  \\
    MAGVLT$_\text{Large}$ & \bf{55.8} & \bf{9.8} \\ 
    \hline
    \end{tabular}
}
\caption{\textit{Zero-shot} I2T results on MS-COCO Karpathy test ({\bf Top}) and NoCaps validation ({\bf Bottom}). \label{tab:scaling_i2t}}
\end{table}
\egroup

The zero-shot I2T results on MS-COCO and NoCaps datasets are presented in \cref{tab:scaling_i2t}. Similar to the T2I results, the large-scale models of both VLTs show better I2T scores compared to their respective default models. Note that in case of ARGVLT, the performance gap between the default and large models is marginal on MS-COCO dataset, while MAGVLT improves the performance significantly on both datasets, as the model size is increased. These results imply that our MAGVLT is more effective in model scaling.  \\

\section{Finetuning on Downstream Tasks}
In order to verify the transferability of MAGVLT by task-specific finetuning, we perform finetuning on two downstream tasks, one for generation and the other for understanding.
In this finetuning setting, ARGVLT and MAGVLT are initialized from their 40K pretrained checkpoint, and MAGVLT$_\text{Large}$ is initialized from 60K pretrained checkpoint.
\\[5pt]
\noindent{\bf{Image Captioning.}}
We finetune ARGVLT and MAGVLT on the image caption generation task of MS-COCO 2014 dataset. In specific, we finetune the VLTs with the cross entropy loss for 100 epochs with a batch size of 512.
The learning rate is set to $10^{-5}$ for ARGVLT and MAGVLT, and $2 \times 10^{-5}$ for MAGVLT$_\text{Large}$.
Note that we do not use the additional tasks, UnrollMask and MixSel, in finetuning.
The captioning performances are presented in \cref{tab:ds_caption}.
Similar to zero-shot I2T results, MAGVLT shows better results compared to ARGVLT. Moreover, the large-scale model of MAGVLT improves the performances compared to its respective default model.
\bgroup
\def\arraystretch{1.05}%
\begin{table}[h]
\centering
\small
\scalebox{1}{
    \begin{tabular}{lcccc}
    \hline
    \multicolumn{1}{l}{Model} & B-4  & M & C  & S
    \\ \hline
    ARGVLT  & 28.6 & 25.2 & 94.7 & 18.1 \\
    MAGVLT  & 29.3 & 27.1 & 103.3 & 20.5 \\
    MAGVLT$_\text{Large}$  & \bf{32.3} & \bf{27.9} & \bf{110.7} & \bf{21.0} \\ 
    \hline 
    \end{tabular}
}
\caption{Comparisons of finetuned models on MS-COCO Karpathy splits. \label{tab:ds_caption}}
\end{table}
\egroup
\\[1pt]
\noindent{\bf{Visual Question Answering.}}
Masked pretraining is well known as a good representation learning approach for VL \emph{understanding} tasks. Therefore, even though we use a variable mask ratio rather than a low fixed ratio during training for obtaining generation capability of MAGVLT, we can also evaluate the transferability of MAGVLT on a discriminative task. For this, we perform experiments on visual question answering (VQA) task, which is a VL understanding task that requires a model to answer a question given an image, on the commonly used VQAv2 dataset \cite{vqa2017}.
Following \cite{wang2021simvlm}, we treat this task as a classification task where an auxiliary classifier predicts an answer from 3,129 candidates.
The tokens of the question mark `{\fontfamily{qcr}\selectfont ?}' and {\fontfamily{qcr}\selectfont <MASK>} token are sequentially added to the tail of the input sequence $[X; Y]$ where $[\cdot]$ is the concatenation operator.
The top layer output of {\fontfamily{qcr}\selectfont <MASK>} is used as an input for the classifier.
We finetune the classifier and the corresponding model with the cross entropy loss for 20 epochs with a batch size of 2,048 and a learning rate of $5 \times 10^{-5}$, and the dropout rate of the top layer output is set to 0.6.

The results are shown in \cref{tab:ds_vqa}. Compared to the latest algorithms \cite{hao2022language, dai2022enabling}, MAGVLT performs slightly worse, however it can be confirmed that the discriminative representation for understanding has been learned by MAGVLT to some extent. While VLKD \cite{dai2022enabling} and MetaLM \cite{hao2022language} use large-scale language-only data and leverage a language model, we pretrain our model from scratch using only paired image-text datasets. And, our model is basically trained for generation, and moreover, it can even generate images by a single model. 
\bgroup
\def\arraystretch{1.05}%
\begin{table}[h]
\centering
\small
\scalebox{1}{
    \begin{tabular}{lcc}
    \hline
    \multicolumn{1}{l}{Model} & test-dev & test-std
    \\ \hline
    VLKD$_{\text{ViT-B/16}}$ \cite{dai2022enabling} & 69.8 & - \\
    MetaLM \cite{hao2022language} & \bf{74.4} & \bf{74.5} \\
    \hline
    MAGVLT  & 63.0 & 63.4 \\
    MAGVLT$_\text{Large}$ & 65.7 & 66.2 \\ 
    \hline
    \end{tabular}
}
\caption{Experimental results on VQAv2. \label{tab:ds_vqa}}
\end{table}
\egroup
\section{Unconditional Image+Text Generation Result}

\begin{figure*}[t]
\centering
\includegraphics[width=.65\textwidth]{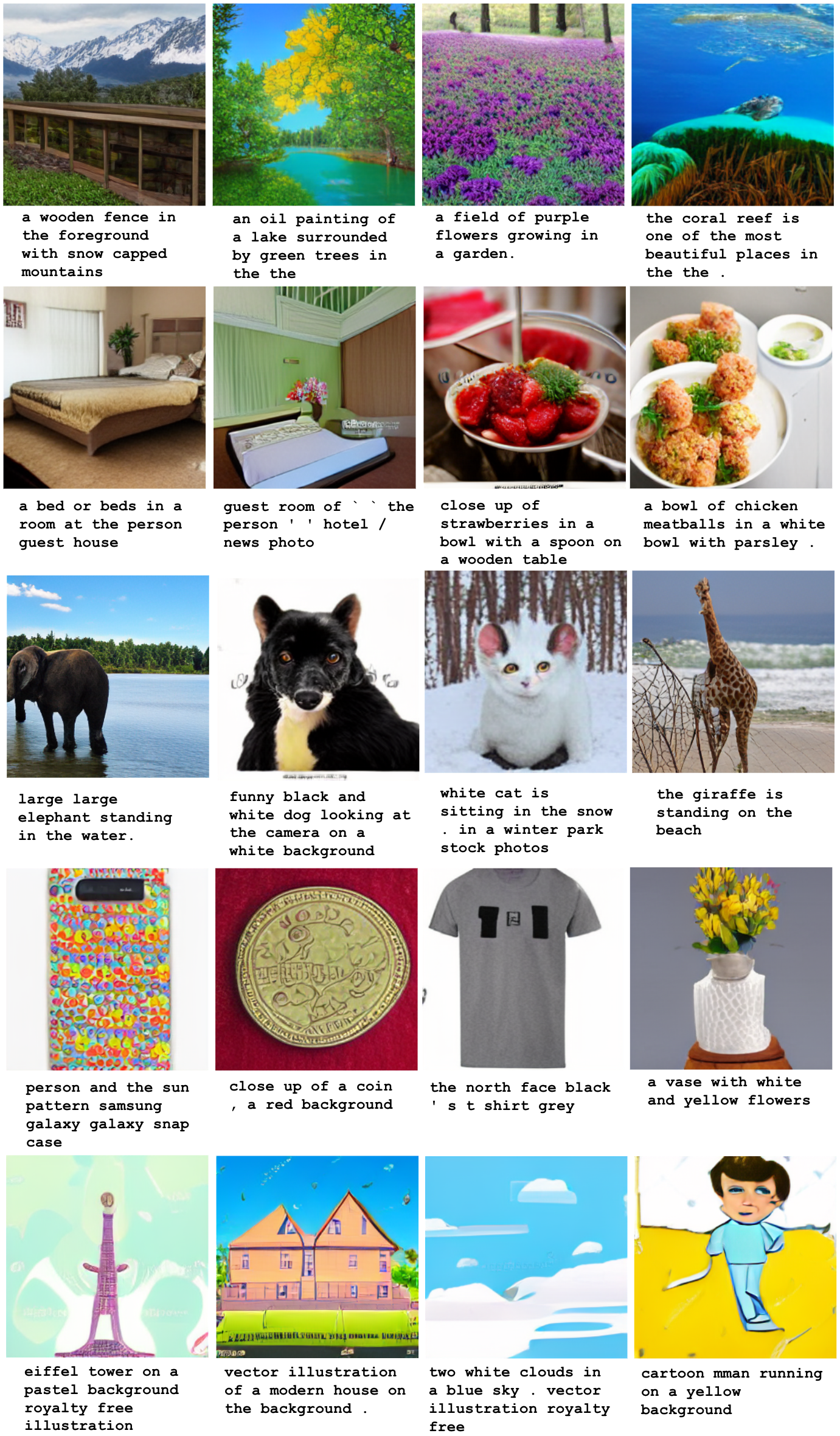}
\caption{\label{fig:unconditioanl_ti} Unconditional image+text generation results obtained by MAGVLT. Note that the generated images cover diverse categories, such as natural scenery (1st row), indoor scenes \& foods (2nd row), animals (3rd row), objects (4th row), and illustrations (5th row). Also, the generated texts are well aligned with generated images.}
\vspace{-0.5cm}
\end{figure*}

Since we train MAGVLT with the three multi-modal tasks including IT2IT, the model is able to produce both image and text at a time. 
Namely, all of the tokens of $X$ and $Y$ are masked at first, and then refined through the iterative decoding. For the target length prediction, the target length is randomly initialized in a range from 8 to 16 and then iteratively predicted as the refinement step proceeds. 
Here, we provide unconditional image+text generation results which are presented in \cref{fig:unconditioanl_ti}. Note that the generated images are very diverse and generally have high quality, and the generated texts also describe the images properly. 
\section{MixSel Analysis}

Here, we demonstrate the effectiveness of the proposed \textit{MixSel} task.
As described in \cref{sec:mixsel}, MixSel mixes two different contexts and selects one of them to be used for generation. 
We hypothesize that our MixSel training task allows the model to attend more carefully to the proper cross-modal context and accordingly to reduce the overlooking of the cross-modal context.
In order to verify this, we first consider \textit{MixRandom} setting which is the same as MixSel, but different in that the target is randomly selected without the additional special token to inform which one is selected, \ie {\fontfamily{qcr}\selectfont <LEFT>} and {\fontfamily{qcr}\selectfont <RIGHT>} or {\fontfamily{qcr}\selectfont <TOP>} and {\fontfamily{qcr}\selectfont <BOTTOM>}. This MixRandom can be seen as the perturbation of the input context alone for regularization like data augmentations.
In \cref{tab:aug}, MAGVLT$_\text{MixRandom}$, which indicates the trained MAGVLT along with UnrollMask and MixRandom, deteriorates the performances of both the zero-shot I2T and the zero-shot T2I, in comparison to MAGVLT with the use of MixSel training.
\bgroup
\def\arraystretch{1.05}%
    \begin{table}[h]
    \centering
    \small
    \begin{tabular}{lcc}
    \hline
    Model & CIDEr ($\uparrow$) & FID ($\downarrow$) \\
    \hline
    MAGVLT$_\text{MixSel}$  & \bf{60.4} & \bf{12.08} \\
    MAGVLT$_\text{MixRandom}$ & 57.9 & 13.43  \\
    \hline
    \end{tabular}
    \caption{Comparison of MixSel and MixRandom on \textit{Zero-shot} I2T and T2I. \label{tab:aug}}
    \end{table}
\egroup

\begin{figure*}[t]
\centering
\includegraphics[width=.9\textwidth]{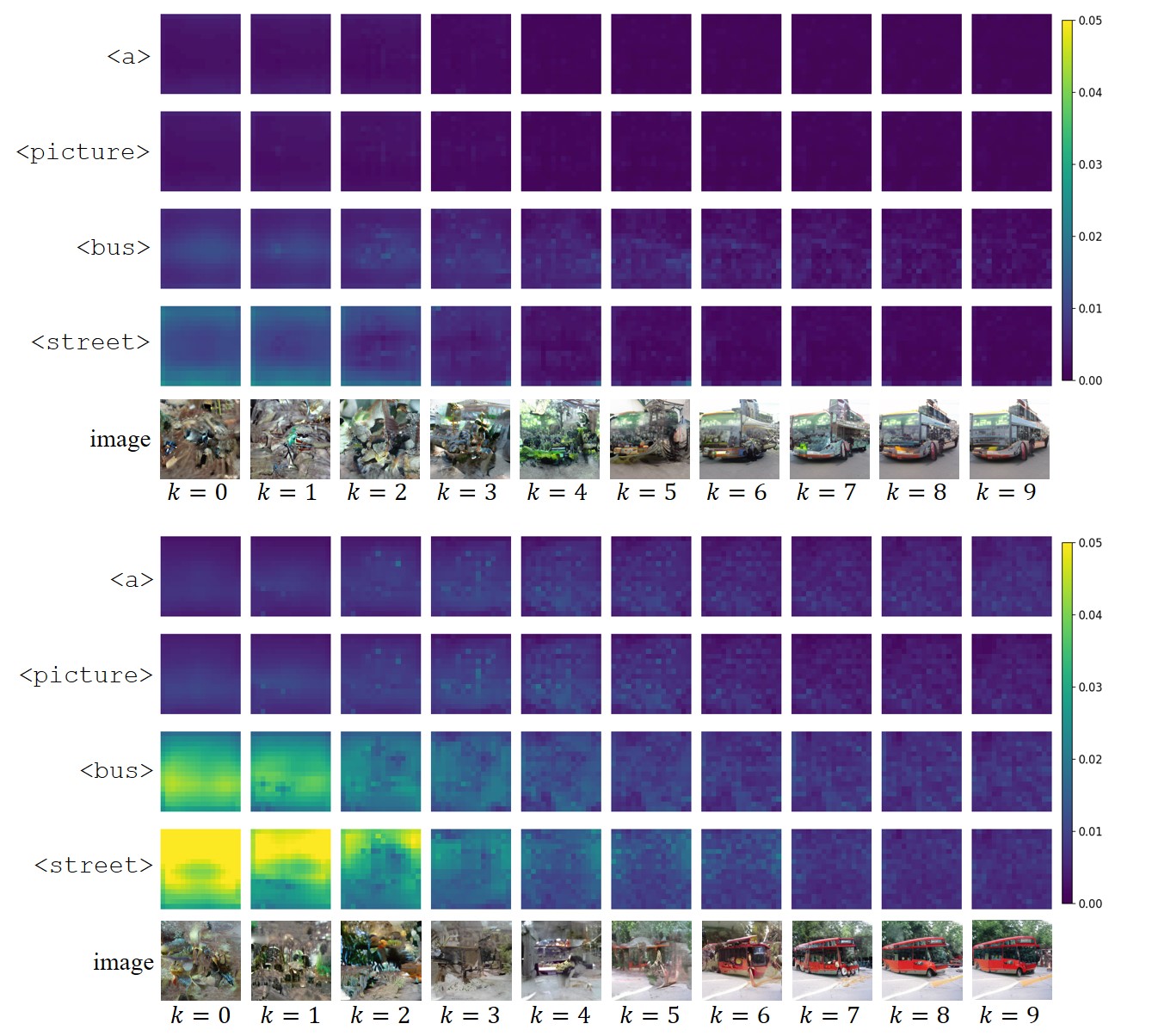}
\caption{\label{fig:mixsel_attn} Visualization of cross-modal attention maps and generated images at different refinement steps. Given the text "a picture of bus in the street.", images are generated using MAGVLTs trained without the use of UnrollMask and MixSel (\textbf{Top}) and with the use of UnrollMask and MixSel (\textbf{Bottom}). To visualize each attention map, cross-attention scores between all 256 image tokens (queries) and a specific text token (a key, corresponds to each row) are computed and then reshaped to 16x16. Image tokens more attend to object text tokens ({\fontfamily{qcr}\selectfont <bus>} and {\fontfamily{qcr}\selectfont <street>}) when the model trained with the use of UnrollMask and MixSel.}
\vspace{-0.5cm}
\end{figure*}

Furthermore, in \cref{fig:mixsel_attn}, we qualitatively show by visualization of cross-modal attention maps that MixSel pretraining task makes the model to attend more to the cross-modal context appropriately compared to the model trained without MixSel training.

\section{Additional Samples}
\begin{figure*}[t]
\centering
\includegraphics[width=.6\textwidth]{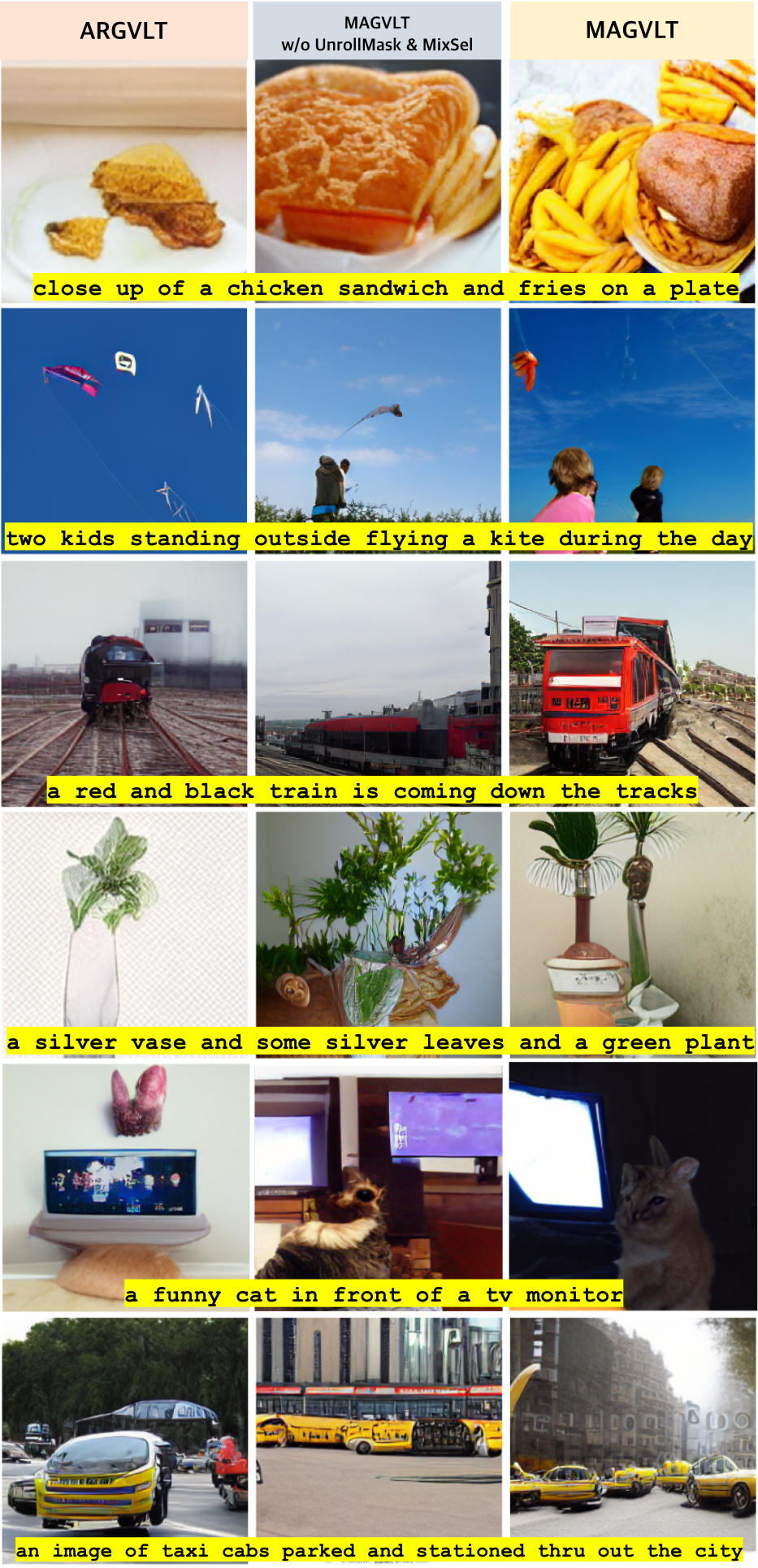}
\caption{\label{fig:t2i_more} More samples of text to image on MS-COCO dataset. }
\vspace{-0.5cm}
\end{figure*}

\begin{figure*}[t]
\centering
\includegraphics[width=.5\textwidth]{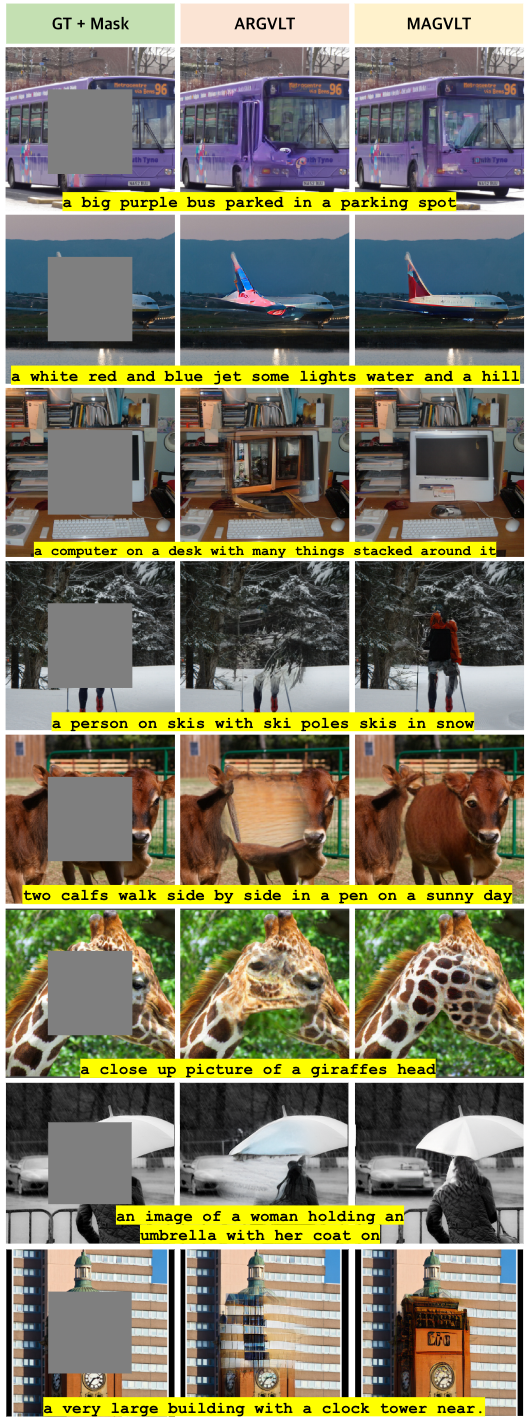}
\caption{\label{fig:t2i_inpaint_more} Image inpainting samples on MS-COCO dataset. MAGVLT generated the masked parts to be more blended with the surrounding context, and more proper to the captions.}
\vspace{-0.5cm}
\end{figure*}

\begin{figure*}[t]
\centering
\includegraphics[width=.85\textwidth]{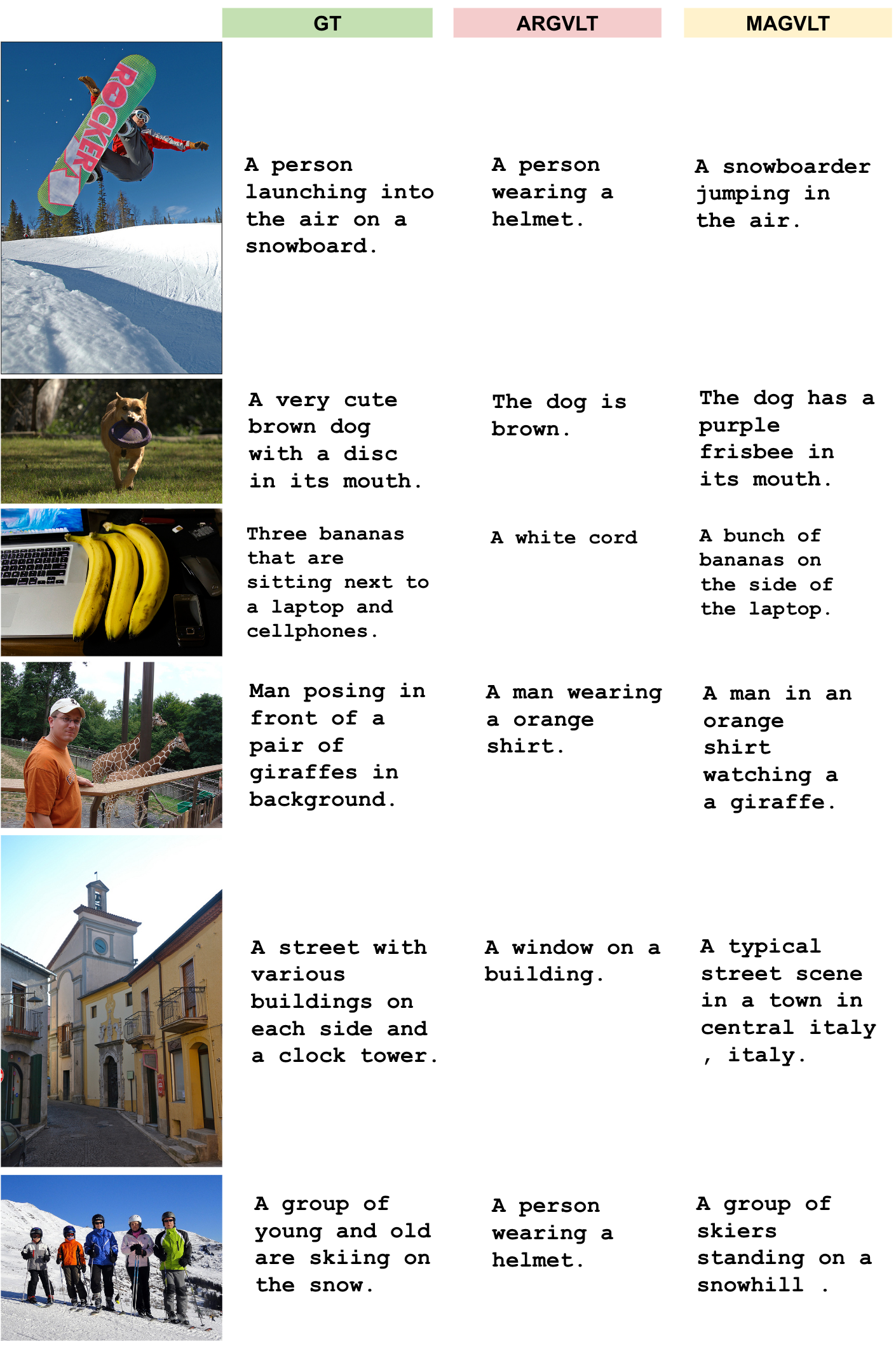}
\caption{\label{fig:i2t_more} More samples of image captioning on MS-COCO dataset. }
\vspace{-0.5cm}
\end{figure*}

\begin{figure*}[t]
\centering
\includegraphics[width=.7\textwidth]{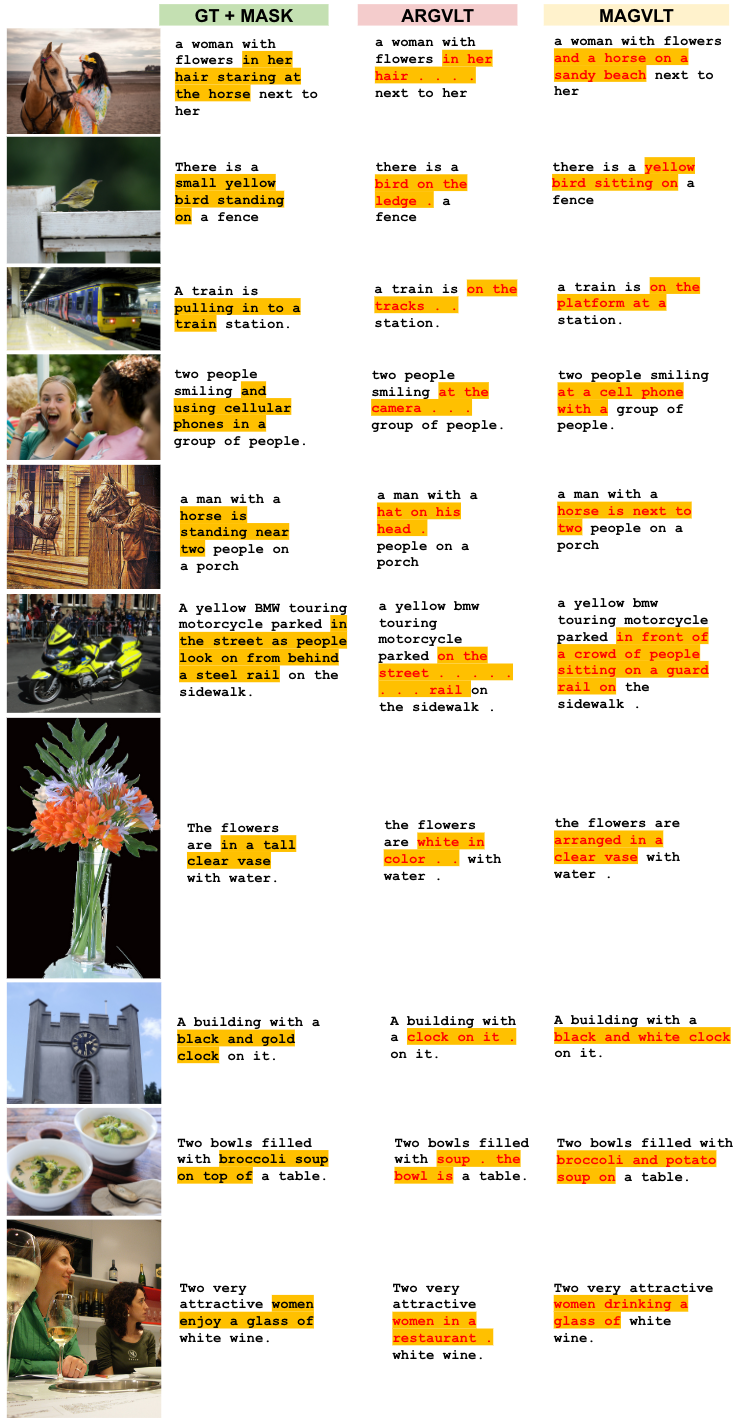}
\caption{\label{fig:i2t_infill_more} Text infilling samples on MS-COCO dataset. The locations to be infilled are shaded with orange color. The words infilled by MAGVLT are better aligned with the surrounding context words, and more appropriate on the corresponding images.}
\vspace{-0.5cm}
\end{figure*}

Here, we present more qualitative results of image and text generation tasks described in \cref{sec:image gen} and \cref{sec:text gen}. The image generation and inpainting results are presented in \cref{fig:t2i_more}, \cref{fig:t2i_inpaint_more}, respectively. The image captioning and text infilling results are shown in \cref{fig:i2t_more} and \cref{fig:i2t_infill_more}, respectively. For text generation tasks, we resize and center-crop the validation images. Overall, our proposed MAGVLT shows better results than ARGVLT.


\end{document}